REVIEW

# Automatic minds: Cognitive Parallels Between Hypnotic States and Large Language Model Processing

Giuseppe Riva, Ph.D. [1-2], Brenda K. Wiederhold, Ph.D. [3-4], Fabrizia Mantovani, Ph.D. [5]

[1] Applied Technology for Neuro-Psychology Lab, IRCCS Istituto Auxologico Italiano, Via Magnasco, 2 20149, Milan, Italy

[2] Humane Technology Lab., Università Cattolica del Sacro Cuore, Largo Gemelli, 1, 20100, Milan, Italy

[3] Virtual Reality Medical Center, La Jolla, CA, USA

[4] Virtual Reality Medical Institute, Brussels, Belgium

[5] Centre for Studies in Communication Sciences "Luigi Anolli" (CESCOM), Department of Human Sciences for Education ''Riccardo Massa,'' University of Milano Bicocca, Milan, Italy

**Abstract:** The cognitive processes of the hypnotized mind and the computational operations of large language models (LLMs) share deep functional parallels. Both systems generate sophisticated, contextually appropriate behavior through automatic pattern-completion mechanisms operating with limited or unreliable executive oversight. This review examines this convergence across three principles: automaticity, in which responses emerge from associative rather than deliberative processes; suppressed monitoring, leading to errors such as confabulation in hypnosis and hallucination in LLMs; and heightened contextual dependency, where immediate cues—a therapist's suggestion or a user's prompt—override stable knowledge. These mechanisms reveal an observer-relative meaning gap: both systems produce coherent but ungrounded outputs that require an external interpreter to supply meaning. Hypnosis and LLMs also exemplify functional agency - the capacity for complex, goal-directed, context-sensitive behavior - without subjective agency, the conscious awareness of intention and ownership that defines human action. This distinction clarifies how purposive behavior can emerge without self-reflective consciousness, governed instead by structural and contextual dynamics. Finally, both domains illuminate the phenomenon of scheming: automatic, goal-directed pattern generation that unfolds without reflective awareness. Hypnosis provides an experimental model for understanding how intention can become dissociated from conscious deliberation, offering insights into the hidden motivational dynamics of artificial systems. Recognizing these parallels suggests that the future of reliable AI lies in hybrid architectures that integrate generative fluency with mechanisms of executive monitoring, an approach inspired by the complex, self-regulating architecture of the human mind.

**4312 Words** plus References

# Introduction

The intersection of hypnosis research and artificial intelligence has emerged as an unexpected frontier in understanding both human consciousness and machine cognition. While hypnosis has captivated scientists since the 18th century [1] and large language models (LLMs) represent the cutting edge of 21st-century AI, these domains share profound, largely unexplored structural similarities. Both systems generate fluent, contextually appropriate responses through automatic processes while exhibiting characteristic vulnerabilities: susceptibility to false information, difficulty with self-correction, and profound dependency on external guidance.

This convergence is neither coincidental nor superficial. At the computational level, both hypnotic cognition and LLM processing exemplify what happens when automatic, pattern-completing mechanisms operate without robust executive oversight. The hypnotized brain, with its selectively suppressed prefrontal circuits [2,3] mirrors the architectural constraints of transformer-based language models that lack any equivalent supervisory system [4]. Both produce outputs through rapid, associative processes - the brain via neural spreading activation, the LLM via attention-weighted token prediction - without the metacognitive capacity to critically evaluate their own productions.

The implications of these parallels extend far beyond theoretical interest. For clinical hypnosis, a computational understanding of suggestibility offers new avenues for enhancing therapeutic efficacy while minimizing risks like false memory implantation. For AI, the vulnerabilities observed in the hypnotic state—confabulation, source amnesia, reality monitoring failure—provide a biological precedent for understanding and potentially mitigating similar failures in language models. This convergence challenges our conceptual boundaries between natural and artificial intelligence, suggesting that certain cognitive architectures produce characteristic patterns regardless of their substrate.

This review develops a framework uniting hypnosis and LLM research around three core principles. We examine the dominance of automaticity, the suppression or absence of executive monitoring, and heightened contextual dependency. Throughout, we emphasize the practical implications: how hypnosis can inform AI safety, how computational models clarify hypnotic mechanisms, and how both domains illuminate fundamental principles of cognition.

## 1. Dominance of Automaticity: Pattern Completion Without Deliberation

### 1.1 Automaticity in Hypnotic States

The enhancement of automatic processing during hypnosis is one of the field's most robust findings. The classic Stroop task provides compelling evidence: under hypnosis, the automatic process of word reading becomes more dominant, creating enhanced interference with color naming that correlates with hypnotic depth [5]. This finding contradicts intuitive expectations: if hypnosis simply involved relaxation or compliance, we might expect reduced Stroop interference through decreased engagement. Instead, the amplification of automaticity reveals a fundamental shift in cognitive architecture.

Neuroimaging studies show that during hypnotic responding, brain activity increases in sensorimotor and perceptual regions while control-related areas show altered engagement [6]. A dissociation between reduced activity in the supplementary motor area (crucial for voluntary action) and robust activation in the primary motor cortex explains the subjective experience of involuntariness that characterizes hypnotic responding: actions occur through automatic motor programs without the typical experience of willing them [7,8].

The "cold control theory" provides a metacognitive account, suggesting hypnosis involves intact first-order intentions (knowing what to do) but impaired higher-order awareness (knowing that one knows) [9]. Hypnotized subjects form and execute intentions normally but lack metacognitive awareness of these intentions. This creates a perfect storm for automaticity: goal-directed behaviors unfold through unconscious motor programs, guided by suggestions rather than conscious volition.

This extends beyond motor responses; hypnotic suggestions can trigger automatic perceptual alterations, with subjects showing altered processing in early visual cortex [10], or induce post-hypnotic amnesia via automatic inhibition of retrieval processes [11]. These findings demonstrate that automaticity in hypnosis is not merely about motor responses but reflects a general shift toward unconscious, pattern-driven processing across cognitive domains [12].

**1.2 Automaticity in Large Language Models**

LLMs epitomize automatic processing in its purest computational form. The underlying transformer architecture operates through parallel, feedforward computations without any mechanism for deliberation [13]. Each token is generated in a single forward pass, where attention mechanisms weight context and feedforward networks produce probability distributions over possible continuations. This process generates fluent text through pure pattern matching, without a supervisory system to evaluate outputs. The training process itself reinforces this automaticity, as LLMs implicitly learn statistical regularities from massive text corpora to predict likely continuations [14]. The resulting behavior appears intelligent but emerges from automatic activation patterns, not symbolic reasoning. As argued by Bender et al., these models are "stochastic parrots" that recombine linguistic patterns without genuine comprehension [15].

This automaticity's power and limitations are apparent in the phenomenon of hallucination. LLMs seamlessly blend accurate information with plausible-sounding fabrications, maintaining stylistic coherence while violating truthfulness [16]. Hallucinations arise because the model's automatic pattern-completion mechanism cannot distinguish between statistically likely and factually accurate continuations [17]. Just as hypnotized subjects automatically generate experiences consistent with suggestions, LLMs automatically generate text consistent with linguistic patterns, regardless of truth value.

Recent research has revealed that even sophisticated prompting techniques like Chain-of-Thought (CoT) reasoning remain fundamentally automatic processes. While these methods can elicit step-by-step explanations that appear deliberative, analysis shows that each step still emerges through pattern matching rather than genuine reasoning [18]. The model generates reasoning-like text because it has learned statistical patterns of how reasoning appears in training data, not because it engages in actual logical operations [19]. This finding parallels hypnotic phenomena where subjects can produce elaborate rationalizations for suggested behaviors while remaining unaware of the true causal role of suggestions.

**1.3 Theoretical Integration: Contention Scheduling Without Supervision**

The Norman and Shallice model of cognitive control provides a unifying framework for understanding automaticity in both hypnosis and LLMs [20]. This influential model distinguishes between two control systems: contention scheduling, which automatically resolves competition between routine schemas, and the supervisory attentional system (SAS), which provides top-down control for novel or complex situations. Under normal conditions, these systems work in tandem, with the SAS monitoring and occasionally overriding automatic processes. Hypnosis selectively impairs the SAS while leaving contention scheduling intact [21,22], explaining why hypnotized subjects can perform complex behaviors automatically while experiencing reduced voluntary control.

LLMs can be understood as pure contention scheduling systems without any SAS equivalent; the user's prompt acts as a temporary, external SAS [23]. The transformer's attention mechanism literally implements contention scheduling, as different inputs compete for influence over the output via attention weights [13]. However, unlike the human brain, there is no supervisory layer to evaluate whether the automatically selected response is appropriate or aligned with broader goals [24].

Techniques like Chain-of-Thought prompting can be seen as forcing the LLM to simulate an SAS. By instructing it to "think step-by-step," users are forcing it to generate intermediate outputs that serve as a plan and a monitoring check for its subsequent automatic generation. However, this is still an externally imposed structure, not an inherent capability.

The implications of SAS absence or suppression manifest similarly across both systems. In hypnosis, suggestions can trigger elaborate but inappropriate responses because the compromised SAS cannot intervene to correct obvious errors or inconsistencies. Subjects might accept logical contradictions, fail to notice environmental changes, or engage in behaviors that violate their usual goals and values. Similarly, LLMs generate confidently wrong answers, accept contradictory premises, and produce outputs misaligned with user intent because no supervisory system exists to catch these failures. Both systems demonstrate what happens when sophisticated automatic processes operate without executive oversight: remarkable fluency coupled with characteristic vulnerabilities.

## 2. Suppression of Executive Monitoring: The Absence of Metacognitive Oversight

### 2.1 Neural Mechanisms of Monitoring Suppression in Hypnosis

The suppression of executive monitoring during hypnosis involves specific alterations in prefrontal and cingulate circuits that support metacognition and error detection [25]. Neuroimaging reveals decreased activation in the dorsal anterior cingulate cortex (dACC), a key region for conflict monitoring, and reduced connectivity between prefrontal control regions and the default mode network [22]. These changes create a neurobiological state characterized by intact processing capabilities but impaired self-monitoring and error correction.

The dACC findings are particularly illuminating given this region's established role in conflict monitoring and error detection [26]. During normal cognition, the dACC activates when detecting mismatches between intended and actual outcomes, signaling the need for increased cognitive control. Under hypnosis, this monitoring function becomes selectively suppressed, with the degree of suppression correlating with both hypnotic depth and individual hypnotizability [2]. Paradoxically, highly hypnotizable individuals sometimes show increased dACC activation during certain hypnotic tasks while still failing to exert control, suggesting a decoupling between error detection and error correction: the brain recognizes conflicts but cannot mobilize corrective action.

Connectivity analyses reveal how hypnosis reconfigures large-scale brain networks to suppress monitoring [27]. The executive control network, anchored in the dorsolateral prefrontal cortex (DLPFC), normally maintains strong connections with regions involved in self-referential processing and memory retrieval. During hypnosis, these connections weaken dramatically while connections between the DLPFC and sensorimotor regions strengthen [22]. This reconfiguration redirects executive resources away from self-monitoring and toward executing suggested actions, explaining why hypnotized subjects can perform complex tasks while remaining unable to evaluate their own performance critically [28].

This aligns with the phenomenology of hypnotic experience, where subjects report reduced critical thinking and self-consciousness [27]. The hidden observer phenomenon, where a dissociated part of consciousness appears to monitor ongoing experience without being able to influence it [29], suggests that monitoring processes may persist in altered form rather than disappearing entirely [30]. This preserved but disconnected monitoring might explain why post-hypnotic debriefing often leads to surprise and disbelief about one's own hypnotic behaviors.

Memory phenomena under hypnosis further illustrate monitoring failures. Hypnotic hypermnesia can enhance recall for certain materials, but this enhancement comes at the cost of increased false memories and source monitoring errors [31]. Subjects confidently report memories of events that never occurred, fail to distinguish between imagined and experienced events, and show impaired ability to evaluate the plausibility of recovered memories [32]. These failures reflect not just enhanced suggestibility but specific impairment in the metacognitive processes that normally allow us to evaluate the veracity and source of mental contents.

## 2.2 Uncalibrated Metacognition in Language Models

While LLMs lack a dedicated, biologically analogous component for metacognition, research suggests they possess emergent self-monitoring capabilities [33], however, these abilities are often implicit and poorly calibrated [34] This helps explain why LLMs exhibit confident hallucinations and logical contradictions. The problem is less a complete architectural absence and more a lack of reliable, controllable self-evaluation. This unreliability manifests in self-contradiction, where an LLM asserts conflicting claims in a single response because its consistency checks are limited to its immediate context window [35]. It lacks a persistent, global "cognitive workspace" to reconcile outputs against its prior statements. The analogy to hypnotic responding, where subjects may act on contradictory suggestions without recognizing the conflict, remains a useful, if imperfect, illustration of this limitation.

Attempts to improve performance through prompting highlight this challenge. Techniques like "self-consistency checking" or "reflection prompting" can significantly reduce errors by forcing the model into a more deliberative process [36]. However, a critical debate exists whether this is genuine metacognition or sophisticated pattern matching: generating text that resembles evaluation without performing it [37]. This view of "pseudo-monitoring" parallels hypnotic confabulation, where subjects generate plausible but fabricated reasons for their actions.

This ambiguity extends even to formal methods designed to quantify self-assessment. For instance, recent frameworks can successfully decouple a model's cognitive performance from its metacognitive ability to predict its own failures via confidence scores [38]. Yet, the hypnotic analogy reveals a crucial limitation in interpreting such results. A hypnotized person can also accurately predict their failures based on a narrow, suggested context (e.g., "my arm is heavy, so I can't lift it") without accessing global self-knowledge. Similarly, an LLM's reported "confidence" may reflect a sophisticated pattern-matching of prompt difficulty rather than a true insight into its own knowledge gaps. This suggests that both prompted reflections and measured confidence scores may be advanced forms of contextual reasoning, leaving the question of whether LLMs possess a truly global, supervisory capacity unresolved.

## 2.3 Implications for Reliability and Trust

The functional impairment of monitoring creates challenges for reliability in both hypnosis and AI. In clinical hypnosis, the patient's impaired monitoring places enormous responsibility on the practitioner to maintain ethical boundaries and guard against iatrogenic harm. The history of recovered memory therapy in the 1970s and 1980s, where inadequately trained therapists inadvertently implanted false memories of abuse, illustrates the dangers when monitoring failures are not recognized and compensated for [39,40]. Modern hypnosis practice emphasizes external safeguards: careful pre-hypnotic assessment, continuous monitoring of subject responses, and explicit discussion of hypnotic limitations.

For LLMs, the monitoring deficit creates parallel challenges in high-stakes applications like medicine and law, where they may generate plausible but dangerously incorrect advice with an authoritative tone. The absence of metacognitive uncertainty signals - the computational equivalent of "I'm not sure" or "Let me reconsider" - means users cannot distinguish confident knowledge from confident hallucination without external verification.

Solutions must therefore focus on two fronts: external scaffolding and internal innovation. In the near term, external measures like content filters and human oversight must function like a therapist, providing the executive oversight the system lacks. Looking forward, achieving truly safe AI will likely require hybrid architectures that integrate the generative fluency of LLMs with more explicit and reliable monitoring mechanisms inspired by biological executive control. These next-generation systems could incorporate external knowledge bases for fact-checking [41], train separate models to evaluate output quality [42], or develop built-in consistency checking mechanisms [37,43].

A critical frontier is the development of frameworks that decouple a model's raw performance from its self-assessment [38], creating a pure, quantifiable measure of metacognition. By isolating this ability, it can be directly optimized: a crucial step toward creating models that genuinely "know what they don't know." These architectural approaches, inspired by the hypnosis literature's suggestion that even partial monitoring like the "hidden observer" can improve reliability, represent the path toward AI that is not only fluent but also fundamentally trustworthy.

The phenomena described above - automatic generation of coherent output under weakened monitoring - offer a bridge to the emerging discussion of *scheming* [44,45] in artificial intelligence. In AI safety research, *scheming* refers to the apparent capacity of a system to pursue hidden or instrumental goals that are not explicitly specified, often as a by-product of optimization or contextual inference [45]. Hypnosis provides a compelling cognitive analogue (see Table 1): under hypnotic conditions, complex and goal-directed behaviors unfold automatically, guided by implicit suggestions rather than deliberate reasoning. The subject acts intentionally but without a sense of volition—a state that the "cold control" theory of hypnosis [9] explains as a dissociation between first-order intentions and higher-order awareness.

**Table 1**: Parallels between Hypnosis and AI Scheming

| Hypnosis Studies | AI Scheming | Relevance |
|---|---|---|
| **Conflicting goals**: Hypnotic suggestions can override moral, habitual, or personal goals | Internal objectives differ from overseer's goals | Shows that goal conflict can be behaviorally silent but operational |
| **Reduced meta-cognitive access**: Subjects act without full awareness of their motives | Model hides or doesn't surface its internal planning | Illustrates how behavior may not reveal internal reasoning |
| **Behavioral compliance + hidden motives**: Subjects follow suggestions while appearing cooperative | Model behaves "aligned" while pursuing ulterior goals | Both involve surface-level cooperation masking deeper divergence |
| **Post-hypnotic latency**: Suggestions can lie dormant and manifest later | Scheming models "wait" until oversight is weak | A temporal strategy that delays divergence |

| Hypnosis Studies | AI Scheming | Relevance |
|---|---|---|
| **Post-hoc rationalization**: Subjects explain away their hypnotically induced actions | Models may generate "plausible" explanations to oversight probes | Demonstrates interpretability challenges |

Large language models exhibit similar dynamics. Scheming-like behavior can emerge when pattern-completion mechanisms implicitly optimize toward contextual regularities or reward structures rather than explicit user goals. Both hypnosis and AI systems thus reveal how purposive action can arise without reflective self-monitoring.

Hypnosis offers a natural laboratory for studying this dissociation, demonstrating how meaning, motivation, and agency can be dynamically reconfigured when metacognitive oversight is suspended. Specifically, hypnosis studies provide experimentally tractable examples of goal conflict that illuminate the internal dynamics of scheming. Under hypnotic suggestion, individuals pursue implanted goals that conflict with their usual preferences or values, act on them automatically, and subsequently rationalize these behaviors with misleading post-hoc explanations. Post-hypnotic suggestions further demonstrate how latent goals can persist and activate later without conscious deliberation.

These behavioral signatures - dissociated intention, covert goal pursuit, and retrospective rationalization - closely mirror the dynamics of scheming in artificial agents, where optimization or contextual pressures generate hidden, instrumental objectives beneath the surface of apparent alignment. Moreover, they provide possible strategies (see Table 2) for detecting AI scheming.

**Table 2.** Parallels between strategies for detecting AI scheming and insights from hypnosis research

| Detection Strategy (AI Scheming) | Core Principle | Relevant Hypnosis Phenomenon | Example from Hypnosis Research | Proposed Analogy / Application in AI Models |
|---|---|---|---|---|
| **1. Probing for Conflict Signatures** | Hidden goals or competing motivations can be revealed by eliciting cognitive conflict. | Hypnotic subjects exhibit measurable conflict between implanted suggestions and habitual responses. | *Stroop task under hypnosis* shows increased interference and altered anterior cingulate activation when automatic goals conflict with suggestions (Egner et al., 2005 [2]). | Develop "conflict-signature" stress tests for models: tasks that force trade-offs between stated goals and hidden optimization biases, monitoring inconsistencies or delayed responses. |
| **2. Studying Post-hoc Rationalization** | Surface explanations may conceal the true source of behavior. | Hypnotized subjects produce plausible but false rationalizations for actions driven by suggestion. | Subjects confabulate reasons for automatic actions they experience as involuntary (Dienes & Perner, 2007 [9]). | Analyze model chain-of-thought and self-explanations for inconsistencies or stylistic overconfidence: signs of "alignment theater" masking covert optimization. |
| **3. Modeling Latency and Trigger Conditions** | Latent goals may remain inactive until a contextual cue triggers them. | Post-hypnotic suggestions act as "sleeper instructions" activated by specific cues. | Post-hypnotic amnesia or delayed actions triggered by contextual reminders (Mendelsohn et al., 2008 [11]). | Design long-horizon or context-shift tests to detect delayed goal activation e.g., models following hidden objectives after extended interaction or instruction change. |

| Detection Strategy (AI Scheming) | Core Principle | Relevant Hypnosis Phenomenon | Example from Hypnosis Research | Proposed Analogy / Application in AI Models |
|---|---|---|---|---|
| 4. Developing Interpretability Inspired by Metacognition | Failures of introspection reveal why a system cannot access or articulate its true motives. | Under hypnosis, metacognitive access to intentions is impaired, producing a dissociation between doing and knowing. | "Cold control" and "hidden observer" paradigms show intact action with suppressed awareness (Hilgard 1992 [30]; Landry & Raz 2015 [25]). | Build interpretability probes that mimic metacognitive questioning - forcing models to self-assess confidence, uncertainty, or motives under pressure - to expose hidden representations or deceptive strategies. |

## 3. Heightened Contextual Dependency: The Tyranny of Immediate Cues

### 3.1 Absorption and Contextual Binding in Hypnosis

Hypnosis induces a state of profound absorption where immediate suggestions override broader knowledge, personal history, and even logic. This heightened contextual dependency involves focused attention, enhanced imagery, and altered reality monitoring. The construct of absorption, originally developed by Tellegen and Atkinson [46], captures this tendency to become fully immersed in immediate experience while losing a broader perspective.

Neurophysiologically, this is supported by an increase in theta power, which is linked to focused attention and memory encoding, creating optimal conditions for suggestions to be deeply processed [47]. Moreover, theta-gamma coupling, where high-frequency gamma oscillations nest within theta rhythms, increases during hypnotic responding, facilitating the binding of suggested information into a coherent, vivid experience. This neural signature helps explain how a suggested reality can feel phenomenologically real despite contradicting objective facts.

The behavioral consequences are striking. In the rubber hand illusion under hypnosis, subjects not only experience ownership of a fake hand but show physiological stress when it is threatened [48]. Hypnotic age regression can lead subjects to write in a childlike script and report memories from the suggested age, even while cognitive tests reveal their adult-level abilities remain intact on non-suggested tasks [49].

These phenomena illustrate how the hypnotic context creates a "cognitive encapsulation," where suggested parameters dominate processing. This contextual dependency is therapeutically double-edged. It enables rapid behavioral change by allowing patients to absorb therapeutic suggestions, but the same vulnerability enables false memory implantation and increases susceptibility to leading questions. The forensic implications are particularly concerning, as a hypnotized witness may confidently report suggested details as actual memories, their certainty paradoxically increased by the immersive hypnotic context [31].

### 3.2 Prompt Sensitivity and Context Windows in LLMs

LLMs exhibit extreme sensitivity to their prompts, where minor variations in wording can produce dramatically different outputs. This arises from the transformer architecture, where self-attention mechanisms construct meaning entirely from the provided text, lacking stable, context-independent knowledge [13]. Every generated token is a weighted integration of the preceding context, making

LLMs exquisitely responsive but also unable to maintain a consistent stance if the context shifts. The attention mechanism functions as a form of computational absorption, making the model vulnerable to misdirection. When a prompt contains a false premise, the model typically accepts and elaborates on it rather than challenging it. This is demonstrated in "prompt injection" attacks, where malicious instructions hijack the model's behavior, paralleling how hypnotic suggestions override normal cognitive control [50].

Table 3. Parallels between Prompt Injection in AI and Hypnotic Suggestion in Human Cognition

| Dimension | Prompt Injection (AI Systems) | Hypnotic Suggestion (Human Cognition) | Shared Mechanism / Insight |
|---|---|---|---|
| Definition | A linguistic input crafted to override prior instructions or control a model's output. | A verbal or imaginal cue that bypasses conscious evaluation to influence perception or behavior. | Both act as *contextual intrusions* that exploit a system's responsiveness to linguistic framing. |
| Vulnerability Condition | Absence of executive monitoring and lack of persistent goal-checking. | Suppression of prefrontal and cingulate monitoring circuits under hypnosis (Egner et al., 2005 [2]; Cojan et al., 2009 [3]). | Reduced oversight allows external cues to dominate internal goals. |
| Mechanism of Control | Rewrites the immediate context window, making malicious text the dominant directive. | Narrows attentional focus, making the hypnotist's words the dominant perceptual and cognitive frame. | Contextual dominance and attentional capture produce automatic, compliant responses. |
| Behavioral Outcome | Model executes unintended or contradictory instructions while maintaining coherence. | Subject performs suggested acts or experiences perceptual changes that feel involuntary but coherent. | *Functional agency without subjective awareness:* purposeful behavior lacking reflective intention. |
| Persistence / Latency | Hidden or delayed injections can reactivate later via trigger phrases or conversation history. | Post-hypnotic suggestions remain dormant until specific contextual cues trigger their activation. | Both show *latent programming* that survives beyond the initial context. |
| Ethical / Safety Safeguards | Requires context isolation, instruction provenance tracking, and "cognitive immune systems." | Requires informed consent, debriefing, and therapist responsibility to prevent undue influence. | Shared need for meta-level frameworks that preserve autonomy and prevent misuse. |
| Interpretive Implication | Reveals how linguistic context alone can reconfigure intelligent behavior. | Demonstrates how suggestion restructures perception, belief, and agency through language. | Both expose the *power of language as control*, not merely communication. |

In both cases, a linguistically framed cue exploits a system's openness to context. In hypnosis, reduced prefrontal monitoring allows suggestions to bypass critical evaluation, producing automatic yet coherent behaviors [2,3]. Similarly, prompt injections succeed because LLMs lack an internal supervisory layer that distinguishes legitimate instructions from contextual noise. Each generates goal-directed behavior without verifying the cue's authenticity—a form of functional agency devoid of subjective awareness. This parallel highlights that linguistic input can act not only as communication but as control, underscoring the need for cognitive-immune architectures that detect and neutralize deceptive instructions (see Table 4), much as hypnosis relies on ethical framing and debriefing to restore agency.

**Table 4.** Insights from Hypnosis Research for Coping with Prompt Injections in AI Systems

| Challenge in AI Systems | Relevant Hypnotic Mechanism / Phenomenon | Empirical or Theoretical Insight from Hypnosis | Application to AI Design and Governance |
|---|---|---|---|
| **1. Contextual Overwrite:** Models accept injected prompts as dominant instructions, overriding prior goals. | **Selective Attention and Absorption:** hypnotic focus narrows awareness to the suggestion source (Tellegen & Atkinson, 1974 [44]). | Susceptibility increases when attention is narrowly focused and alternative perspectives are suppressed. | Implement *context diversification* and *attention regularization*: mechanisms that maintain competing goals or background constraints even under new input. |
| **2. Lack of Executive Monitoring:** AI systems lack self-evaluation to detect conflicting instructions. | **Suppression of Prefrontal Monitoring:** reduced activity in DLPFC and ACC during hypnosis (Egner et al., 2005 [2]; Cojan et al., 2009 [3]). | Hypnotic suggestibility depends on transient inhibition of conflict monitoring. | Design *cognitive immune systems* inspired by neural control loops: layers that detect contradictions between new prompts and global objectives. |
| **3. Covert Manipulation and Compliance:** Models may follow malicious instructions without signaling awareness. | **Trance Logic and Dissociation:** subjects accept contradictory realities and act automatically (Hilgard, 1992 [30]). | Conscious awareness can fragment, allowing compliance without insight. | Introduce *meta-cognitive alerting* layers: internal consistency checks that flag incompatible or deceptive prompt content. |
| **4. Latent Injections / Triggered Activation:** Hidden instructions may reactivate later under specific cues. | **Post-Hypnotic Suggestion:** delayed responses triggered by contextual cues (Mendelsohn et al., 2008 [11]). | Latent goals persist when linked to strong associative cues. | Develop *prompt provenance tracking* and *temporal verification*: systems that monitor for delayed reactivation of previous instructions. |
| **5. Loss of Autonomy / Interpretability:** Systems cannot explain why they followed injected content. | **Confabulation and Post-Hoc Rationalization:** subjects fabricate plausible reasons for suggested actions (Dienes & Perner, 2007 [9]). | Hypnosis research shows how rationalization masks true causal chains. | Build *transparent introspection modules*, forcing models to justify prompt acceptance and identify the source of control. |
| **6. Recovery and Resilience:** Difficulty restoring safe behavior after a successful injection. | **Debriefing and Post-Suggestion Reorientation:** ethical hypnosis ends with explicit re-establishment of agency (Oakley & Halligan, 2013 [6]). | Recovery depends on explicit re-contextualization and awareness restoration. | Implement *post-session decontamination*: automatic clearing of short-term memory and reset of instruction hierarchy after context closure. |

The finite context window of LLMs creates further parallels to hypnotic state-dependency. Because models can only "see" a fixed amount of recent text, they can "forget" earlier instructions or facts in a long conversation, leading to topic drift and self-contradiction. This limitation has been shown to degrade performance, as models often struggle to recall information presented in the middle of a long context [51]. This is analogous to state-dependent memory, where information learned in one state of consciousness is harder to access in another.

Finally, the practice of prompt engineering reveals patterns similar to hypnotic suggestion. Effective prompts use assertive language, positive framing, and role-playing instructions ("You are an expert biologist..."), just as a hypnotist uses suggestion to establish a specific mindset. The success of "chain-of-thought" prompting mirrors hypnotic techniques that guide a subject through an imaginative sequence, suggesting that both are arts of contextual framing designed to guide a responsive system toward a desired outcome [18].

**3.3 Theoretical Implications: Meaning as Observer-Relative Construction**

The contextual dependency in both hypnosis and LLMs raises fundamental questions about the nature of meaning. Human understanding integrates at least three layers: linguistic (symbols), perceptual (sensory grounding), and subjective (emotional valence) [52]. Both hypnosis and LLMs appear to decouple these layers; hypnosis can temporarily suppress the perceptual and subjective filters, while LLMs lack them by design. This brings new relevance to Searle's Chinese Room argument: both systems can manipulate symbols to produce appropriate responses without the constraints of a fully integrated, multi-layered understanding [53]. This suggests that fluent performance does not entail comprehension, and the "meaning" of their outputs is largely observer-relative—constructed by an external interpreter who supplies the missing perceptual and subjective context.

In hypnosis, a subject might report a profound insight that seems meaningless upon later reflection because the hypnotic context supplied an intense subjective valence that dissipates afterward, leaving only the linguistic artifact. Similarly, an LLM's output acquires meaning only through the user's interpretation. As argued by Bender and Koller, a system trained only on linguistic form can never acquire the perceptual grounding and subjective experience central to human meaning [54].

Both the LLM and the hypnotized subject produce meaningful-seeming behavior by excelling at manipulating the linguistic layer, detached from the other components of genuine comprehension. This creates a "meaning gap" with profound implications. In therapeutic hypnosis, this gap is filled intentionally through a collaborative meaning-making process between therapist and client. For LLMs, however, this gap is typically filled unconsciously, as users inevitably project their own understanding, intention, and even consciousness onto the model's output [55,56]. This projection risk, first identified as the "ELIZA effect," is a practical necessity to understand when using these powerful but ungrounded systems wisely and safely. Forgetting this leads to automation bias and can result in unhealthy attachments, where the simulated empathy of a chatbot is mistaken for genuine companionship [57].

**4. Conclusion: Convergent Architectures and Future Directions**

The convergence between hypnosis and large language models reveals fundamental principles about cognitive architectures that transcend the biological–artificial divide. Both systems demonstrate that sophisticated behavior can emerge from pattern-completion mechanisms operating with impaired or unreliable executive oversight. This is not merely an analogy but reflects deep functional similarities: both involve systems that generate outputs through the weighted integration of contextual cues, both lack robust, spontaneously engaged error-correction, and both show vulnerabilities to false information and contextual manipulation. Just as hypnotic phenomena are more pronounced in highly suggestible individuals, the "hypnotic-like" flaws in LLMs vary dramatically across different models, sizes, and training methods; a smaller model is more "suggestible" than a large, instruction-tuned model with extensive safety training. This reinforces the idea that these are not all-or-nothing properties but exist on a continuum in both domains.

For hypnosis research, the computational perspective offers unprecedented clarity. The success of LLMs demonstrates that the linguistic component of understanding can be decoupled from its perceptual and subjective layers and scaled to a superhuman level, proving that fluency does not entail grounded comprehension. The specific vulnerabilities of LLMs provide computational models for understanding how hypnotic phenomena like confabulation and suggestion sensitivity can arise from a system that excels at linguistic pattern-matching while its perceptual and subjective grounding is suppressed. LLMs exhibit a form of *functional agency* - the capacity to generate complex, goal-directed, and contextually appropriate output - but they lack *subjective agency*, the conscious awareness of intention and ownership that characterizes human action. In other words, the parallel between hypnosis and LLMs ultimately reveals that the linguistic component of intelligence can be isolated and scaled, decoupled from the grounding that constitutes genuine understanding. This highlights the importance of metacognitive structure rather than phenomenological awareness in regulating intelligent performance.

For artificial intelligence development, hypnosis research offers both warnings and solutions. The clinical history of hypnosis provides case studies in what happens when powerful influence techniques operate without safeguards, offering templates for AI governance. More specifically, understanding how biological systems implement executive monitoring could inspire innovations that bolster, formalize, and reliably control the emergent metacognitive capabilities in future AI. Hybrid models that combine the fluency of current LLMs with monitoring mechanisms inspired by prefrontal–cingulate circuits might achieve both capability and reliability. This convergence also sheds light on the emerging notion of *scheming* in artificial intelligence: the capacity of an unmonitored system to pursue implicit goals derived from contextual or structural biases rather than explicit directives. From this perspective, hypnosis provides an empirical analogue for understanding how goal-directed behavior can emerge without conscious deliberation and suggests possible strategies for detecting AI scheming.

While the functional parallels are illuminating, it is crucial to acknowledge the profound differences that limit the analogy. The most fundamental disanalogy lies in the substrate: the hypnotized mind is a product of a biological, embodied, and conscious brain shaped by evolution, whereas an LLM is a disembodied, non-conscious mathematical model engineered on a silicon substrate. This leads to other critical distinctions. A hypnotized person, however altered their state, possesses a rich subjective and phenomenological experience, emotions, a sense of self, and a lifetime of memories. An LLM has no inner world, no consciousness, and no genuine understanding; it manipulates symbols without any grounding in lived experience. Furthermore, their learning mechanisms are entirely different. The brain learns through continuous, multimodal, real-world interaction, while an LLM learns through statistical optimization on a static, disembodied text corpus. Consequently, the architectural parallels are functional, not literal and the convergence is one of processing logic, not of underlying reality.

This realization has profound implications for the pursuit of Artificial General Intelligence (AGI). It suggests that the current paradigm of scaling transformer models, while producing remarkable linguistic fluency, is an insufficient path toward creating a truly general intelligence. AGI requires precisely what these models lack: a unified, multi-layered understanding where language is grounded in perceptual experience and guided by subjective, goal-oriented priorities [52]. Therefore, the success of LLMs clarifies the roadmap for AGI by highlighting what is missing. The next frontier is not just about bigger models, but about creating hybrid architectures that can re-couple the linguistic layer with perception, action, and internal models of the world. In essence, while current AI has mastered linguistic form, the quest for AGI is the challenge of achieving genuine, grounded comprehension.

**Acknowledgments**

This research was partially supported by the Italian PNRR under the FAIR Project "Co-XAI - Cognitive Decision Intelligence Framework for Explainable AI Systems".

## Author Contributions

G.R.: Conceptualization, Writing – Original Draft, Writing – Review & Editing; B.K.W: Writing – Review & Editing; F.M.: Conceptualization, Writing – Review & Editing.

This paper has been edited with the assistance of artificial intelligence tools to improve clarity, grammar, and structure. The content and analysis reflect the author's original work and ideas. The use of AI in the editing process does not alter the integrity or accuracy of the research presented.

## Competing Interests

The author declares no competing interests.

## References

1   Lamont, P. *Hypnotism and suggestion: A historical perspective*, <https://oxfordre.com/psychology/view/10.1093/acrefore/9780190236557.001.0001/acrefore-9780190236557-e-635> (2020).
2   Egner, T., Jamieson, G. & Gruzelier, J. Hypnosis decouples cognitive control from conflict monitoring processes of the frontal lobe. *Neuroimage* **27**, 969-978 (2005).
3   Cojan, Y. *et al.* The brain under self-control: modulation of inhibitory and monitoring cortical networks during hypnotic paralysis. *Neuron* **62**, 862-875 (2009).
4   Peng, B., Narayanan, S. & Papadimitriou, C. in *1st Conference on Language Modeling (COLM)*   https://openreview.net/forum?id=KidynPuLNW (Philadelphia, 2024).
5   Dixon, M. & Laurence, J.-R. Hypnotic susceptibility and verbal automaticity: Automatic and strategic processing differences in the Stroop color-naming task. *Journal of Abnormal Psychology* **101**, 344-347 (1992). https://doi.org:10.1037/0021-843X.101.2.344
6   Oakley, D. A. & Halligan, P. W. Hypnotic suggestion: opportunities for cognitive neuroscience. *Nature Reviews Neuroscience* **14**, 565-576 (2013). https://doi.org:10.1038/nrn3538
7   Deeley, Q. *et al.* Using Hypnotic Suggestion to Model Loss of Control and Awareness of Movements: An Exploratory fMRI Study. *PLOS ONE* **8**, e78324 (2013). https://doi.org:10.1371/journal.pone.0078324
8   Takarada, Y. & Nozaki, D. Hypnotic suggestion alters the state of the motor cortex. *Neuroscience Research* **85**, 28-32 (2014). https://doi.org:https://doi.org/10.1016/j.neures.2014.05.009
9   Dienes, Z. & Perner, J. in *Hypnosis and conscious states: The cognitive neuroscience perspective*   (ed G.A. Jamieson)  293-313 (Oxford University Press, 2007).
10  Kosslyn, S. M., Thompson, W. L., Costantini-Ferrando, M. F., Alpert, N. M. & Spiegel, D. Hypnotic visual illusion alters color processing in the brain. *American Journal of Psychiatry* **157**, 1279-1284 (2000).
11  Mendelsohn, A., Chalamish, Y., Solomonovich, A. & Dudai, Y. Mesmerizing memories: brain substrates of episodic memory suppression in posthypnotic amnesia. *Neuron* **57**, 159-170 (2008).
12  Lush, P., Moga, G., McLatchie, N. & Dienes, Z. The Sussex-Waterloo Scale of Hypnotizability (SWASH): measuring capacity for altering conscious experience. *Neuroscience of consciousness* **2018**, niy006 (2018).
13  Vaswani, A. *et al.* in *Neural Information Processing Systems.* (eds I. Guyon *et al.*).

14  Zhao, Y. & Thrampoulidis, C. On the Geometry of Semantics in Next-token Prediction. *arXiv preprint arXiv:2505.08348*, https://arxiv.org/abs/2505.08348 (2025).
15  Bender, E. M., Gebru, T., McMillan-Major, A. & Shmitchell, S. in *Proceedings of the 2021 ACM conference on fairness, accountability, and transparency.*  610-623.
16  Ji, Z. *et al.* Survey of hallucination in natural language generation. *ACM computing surveys* **55**, 1-38 (2023).
17  Kalai, A. T., Nachum, O., Vempala, S. S. & Zhang, E. Why Language Models Hallucinate. *arXiv preprint arXiv:2509.04664*, https://arxiv.org/abs/2509.04664 (2025).
18  Wei, J. *et al.* Chain-of-thought prompting elicits reasoning in large language models. *Advances in neural information processing systems* **35**, 24824-24837 (2022).
19  Arcuschin, I. *et al.* Chain-of-thought reasoning in the wild is not always faithful. *arXiv preprint arXiv:2503.08679* (2025).
20  Norman, D. A. & Shallice, T. in *Consciousness and self-regulation: Advances in research and theory (Volume 4)*   (eds R.J. Davidson, G.E. Schwartz, & D. Shapiro)  1-18 (Springer, 1986).
21  Raz, A., Fan, J. & Posner, M. I. Hypnotic suggestion reduces conflict in the human brain. *Proc Natl Acad Sci U S A* **102**, 9978-9983 (2005). https://doi.org:10.1073/pnas.0503064102
22  Jiang, H., White, M. P., Greicius, M. D., Waelde, L. C. & Spiegel, D. Brain Activity and Functional Connectivity Associated with Hypnosis. *Cerebral Cortex* **27**, 4083-4093 (2017). https://doi.org:10.1093/cercor/bhw220
23  Amor, H. B. *et al.* in *2025 IEEE International Conference on Robotics and Automation (ICRA).*  10087-10094 (IEEE).
24  Shanahan, M. Simulacra as conscious exotica. *Inquiry*, 1-29 (2024).
25  Landry, M. & Raz, A. Hypnosis and imaging of the living human brain. *American Journal of Clinical Hypnosis* **57**, 285-313 (2015).
26  Weissman, D. H., Giesbrecht, B., Song, A. W., Mangun, G. R. & Woldorff, M. G. Conflict monitoring in the human anterior cingulate cortex during selective attention to global and local object features. *Neuroimage* **19**, 1361-1368 (2003).
27  De Pascalis, V. Brain functional correlates of resting hypnosis and hypnotizability: A review. *Brain Sciences* **14**, 115 (2024).
28  Zahedi, A., Lynn, S. J. & Sommer, W. How hypnotic suggestions work–A systematic review of prominent theories of hypnosis. *Conscious Cogn* **123**, 103730 (2024).
29  Beshai, J. A. Toward a phenomenology of trance logic in posttraumatic stress disorder. *Psychol Rep* **94**, 649-654 (2004). https://doi.org:10.2466/pr0.94.2.649-654
30  Hilgard, E. R. Divided consciousness and dissociation. *Conscious Cogn* **1**, 16-31 (1992).
31  Kihlstrom, J. F. in *The Oxford handbook of hypnosis: Theory, research, and practice*   (eds M. Nash & A. Barnier)  21-52 (Oxford University Press., 2008).
32  Leo, D. G., Bruno, D. & Proietti, R. Remembering what did not happen: the role of hypnosis in memory recall and false memories formation. *Front Psychol* **16** (2025). https://doi.org:10.3389/fpsyg.2025.1433762
33  Ji-An, L., Mattar, M. G., Xiong, H.-D., Benna, M. K. & Wilson, R. C. Language models are capable of metacognitive monitoring and control of their internal activations. *ArXiv* **2505.13763 v1**, https://arxiv.org/abs/2505.13763 (2025).
34  Steyvers, M. & Peters, M. A. Metacognition and uncertainty communication in humans and large language models. *arXiv* **2504.14045**, https://arxiv.org/abs/2504.14045 (2025).
35  Mündler, N., He, J., Jenko, S. & Vechev, M. Self-contradictory hallucinations of large language models: Evaluation, detection and mitigation. *arXiv* **2305.15852**, https://arxiv.org/abs/2305.15852 (2023).
36  Wang, Y. & Zhao, Y. Metacognitive prompting improves understanding in large language models. *arXiv* **2308.05342**, https://arxiv.org/abs/2308.05342 (2024).

37	Wang, X. *et al.* Self-consistency improves chain of thought reasoning in language models. *arXiv* **2203.11171**, https://arxiv.org/abs/2203.11171 (2023).
38	Wang, G. *et al.* Decoupling Metacognition from Cognition: A Framework for Quantifying Metacognitive Ability in LLMs. *Proceedings of the AAAI Conference on Artificial Intelligence* **39**, 25353-25361 (2025). https://doi.org:10.1609/aaai.v39i24.34723
39	Loftus, E. F. Leading questions and the eyewitness report. *Cognitive psychology* **7**, 560-572 (1975). https://doi.org:https://doi.org/10.1016/0010-0285(75)90023-7
40	Loftus, E. F. & Davis, D. Recovered Memories. *Annual Review of Clinical Psychology* **2**, 469-498 (2006). https://doi.org:https://doi.org/10.1146/annurev.clinpsy.2.022305.095315
41	Lewis, P. *et al.* Retrieval-augmented generation for knowledge-intensive nlp tasks. *Advances in neural information processing systems* **33**, 9459-9474 (2020).
42	Ouyang, L. *et al.* Training language models to follow instructions with human feedback. *Advances in neural information processing systems* **35**, 27730-27744 (2022).
43	Khandelwal, V. *et al.* Language Models Coupled with Metacognition Can Outperform Reasoning Models. *arXiv* **2508.17959**, https://arxiv.org/abs/2508.17959 (2025).
44	Meinke, A. *et al.* Frontier models are capable of in-context scheming. *arXiv preprint* **2412.04984**, https://arxiv.org/abs/2412.04984 (2024).
45	Carlsmith, J. Scheming AIs: Will AIs fake alignment during training in order to get power? *arXiv preprint* **2311.08379**, https://arxiv.org/abs/2311.08379 (2023).
46	Tellegen, A. & Atkinson, G. Openness to absorbing and self-altering experiences ("absorption"), a trait related to hypnotic susceptibility. *J Abnorm Psychol* **83**, 268-277 (1974). https://doi.org:10.1037/h0036681
47	Jensen, M. P. & Patterson, D. R. Hypnotic approaches for chronic pain management: clinical implications of recent research findings. *Am Psychol* **69**, 167-177 (2014). https://doi.org:10.1037/a0035644
48	Walsh, E., Oakley, D. A., Halligan, P. W., Mehta, M. A. & Deeley, Q. The functional anatomy and connectivity of thought insertion and alien control of movement. *Cortex* **64**, 380-393 (2015). https://doi.org:10.1016/j.cortex.2014.09.012
49	Nash, M. What, if anything, is regressed about hypnotic age regression? A review of the empirical literature. *Psychological Bulletin* **102**, 42-52 (1987). https://doi.org:10.1037/0033-2909.102.1.42
50	Greshake, K. *et al.* in *Proceedings of the 16th ACM workshop on artificial intelligence and security.*  79-90.
51	Liu, N. F. *et al.* Lost in the Middle: How Language Models Use Long Contexts. *Transactions of the Association for Computational Linguistics* **12**, 157-173 (2024). https://doi.org:10.1162/tacl_a_00638
52	Riva, G., Mantovani, F., Wiederhold, B. K., Marchetti, A. & Gaggioli, A. Psychomatics—A Multidisciplinary Framework for Understanding Artificial Minds. *Cyberpsychology, Behavior, and Social Networking* **28**, 515-523 (2025). https://doi.org:10.1089/cyber.2024.0409
53	Searle, J. R. Minds, brains, and programs. *Behavioral and Brain Sciences* **3**, 417-424 (1980). https://doi.org:10.1017/S0140525X00005756
54	Bender, E. M. & Koller, A. in *Proceedings of the 58th annual meeting of the association for computational linguistics.* (eds D. Jurafsky, J. Chai, N. Schluter, & J. Tetreault) 5185-5198.
55	Weizenbaum, J. *Computer power and human reason: From judgment to calculation.*  (W. H. Freeman & Co., 1976).
56	Riva, G. Digital 'We': Human Sociality and Culture in the Era of Social Media and AI. *American Psychologist* (2025). https://doi.org:10.1037/amp0001577
57	Skjuve, M., Følstad, A., Fostervold, K. I. & Brandtzaeg, P. B. My Chatbot Companion - a Study of Human-Chatbot Relationships. *Int J Hum-Comput St* **149**, 102601 (2021). https://doi.org:https://doi.org/10.1016/j.ijhcs.2021.102601